\pdfoutput=1
\documentclass[11pt,onecolumn]{article}
\usepackage{files/cvpr}
\usepackage{amsmath, amssymb, graphicx, xfrac, amsfonts, amsbsy, algorithm, color, subfig, cite, url, multirow}
\usepackage[noend]{algpseudocode}
\usepackage[section]{placeins}

\cvprfinalcopy

\def\Dbf{\mathbf{D}}
\def\phibf{\pmb{\phi}}

\def\Phibf{\mathbf{\Phi}}

\def\d{\mathbf{d}}
\def\dG{\d_\text{G}}
\def\dL{\d_\text{L}}

\def\tbf{\pmb{\theta}}

\def\Ybf{ \pmb{ \mathrm{Y} } }

\def \g { \mathbf{D} }

\def\yphin{\pmb{ \tilde{\mathrm{y}} }^{ \pmb{\phi} }_n }

\def\X {\mathrm{x} }
\def\Xbf{ \pmb{ \X } }

\def\Xphibf{\Xbf^{\phibf}}
\def \gXbf { \g \Xbf }

\def \gxXbf { \g_x \Xbf }
\def \gyXbf { \g_y \Xbf }

\def\Ophi{ \Omega^{\pmb{\phi}} }
\def\Mphin{ M^{\phibf}_n }

\def\O'{ \Omega'}

\def\R{\mathbb{R}}

\def\Tcal{\Theta}

\title{Ground Edge based LIDAR Localization without a Reflectivity Calibration for Autonomous Driving}

\author{Juan~Castorena$^{1}$, and Siddharth~Agarwal$^{1}$
	\thanks{$^{1}$First and Second Authors are with the Autonomous Vehicles group, Ford Motor Company, Dearborn, MI 48124 USA
		{\tt\small jcastore@ford.com}}%
}

\makeatletter
\def\BState{\State\hskip-\ALG@thistlm}
\makeatother

\providecommand{\keywords}[1]{\textbf{\textit{Index terms---}} #1}

\begin{document}
\maketitle

\begin{abstract}
		{\normalfont
In this work we propose an alternative formulation to the problem of ground reflectivity grid based localization involving laser scanned data from multiple LIDARs mounted on autonomous vehicles. The driving idea of our localization formulation is an alternative edge reflectivity grid representation which is invariant to laser source, angle of incidence, range and robot surveying motion. Such property eliminates the need of the post-factory reflectivity calibration whose time requirements are infeasible in mass produced robots/vehicles. Our experiments demonstrate that we can achieve better performance than state of the art on ground reflectivity inference-map based localization at no additional computational burden.
}
\end{abstract}


\keywords{Localization, SLAM, autonomous vehicles, sensor fusion, LIDAR, calibration.}


\section{Introduction}
\label{Sec:Introduction} 

One of the most popular and efficient formulations that have been developed for the perception and interaction of mobile robotics with the environment is occupancy-grids or correspondingly inference-grids \cite{Kortenkamp98}. 
%
These are tessellations of the space into 2-D (or 3-D in the case of volumetric-grids) representations of the environment where each cell in the grid can inform about features or properties of an object occupying that space based on a probabilistic estimate of its state \cite{Elfes89}.
%
The proliferation of inference-maps has been motivated by its effectiveness in solving a variety of robotic perception and navigation strategy problems. 
%
For example, the works in \cite{Konolige99, Levinson10, Wolcott14} have used local inference grid maps for localization while the works in \cite{Thrun98, Kortenkamp98} have used them globally for the problem of path planning and navigation. More recent methods have use them under a learning framework to map the raw sensed data to object tracks via recurrent neural networks \cite{Ondruska13}.

One of the environment properties that has gained popularity in the last decade both from a research and industrial standpoint is the ground reflectivity inference grid commonly used in autonomous vehicles \cite{Guan14, Pu.Rutzinger.Vosselman.Elberink.2011, Chin.Olsen.2014, Levinson10, Levinson.Thrun2007, Wolcott14}. This provides the necessary road profile information useful to enable and facilitate localization, vehicle-road interaction, path planning and decision making \cite{Levinson.Thrun2007}. 
%
Among the most effective sensing strategies for ground perception consist on surveying with one or more mobile robot/vehicles outfitted with position sensors (e.g., GPS, IMU, odometry) and LIDAR with both range and reflectivity measurement capabilities. Here, reflectivity implies a factory calibration against a target of known reflectance has been performed. 
%
When a robot/vehicle is surveying a region, each reflectivity measurement of the ground is associated with a cell in the grid depending on the projection of its corresponding range into the reference frame of the inference-grid. These projected measurements are non-uniformly distributed across the ground depending on the sensor configuration, the scanning pattern and robot/vehicle motion while surveying. Note that if the inference-grid lays in a global reference frame as it is the case in the generation of global maps, a (full) bundle adjustment \cite{Durrant-Whyte.Bailey2006, Bailey.Durrant-Whyte2006} correction of position estimates is typically carried out to correct for point-cloud missalignments.

\begin{figure}[t]
	\centering
	\subfloat[Asphalt: $\mu \approx 25, \sigma \approx 12$, $D_{KL}(P || Q) = 0.38 $b]{ {\includegraphics[width = 8.0 cm]{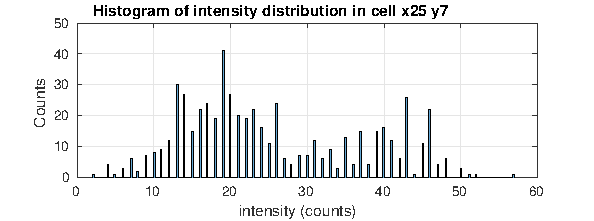} } }
	\subfloat[W. lane: $\mu \approx 30, \sigma \approx 10$, $D_{KL}(P || Q) = 0.31 $b]{ {\includegraphics[width = 8.0 cm]{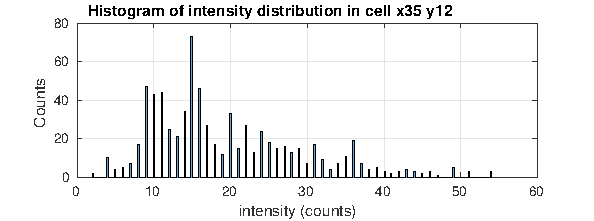}} }	
	\caption{Histogram distributions of reflectivity measurements from multiple HDL-32E's in $10 \times 10$ cm. cells. (a) Cell in plain asphalt. (b) Cell in a white lane road marking.}
	\label{fig:histogramIntensity}
\end{figure}

One of the problems associated with using this additional modality is that its measurements depend on a reflectivity response. This itself is dependent upon the laser-detector (i.e., observer), angle of incidence and range (i.e., perspective) \cite{Kashani15}. Measurements with these dependencies populate the inference-grid where each cell may have measurements from multiple lasers-perspectives and whose number is coupled with the motion of the vehicle at the time of survey, the laser scanning patterns and the laser position configurations. Thus, assuming a reflectivity invariant measurement model is quite often inadequate \cite{Harvey.Lichti02, Burgard.Triebel.Andreasson2005}. 
Na\"{\i}ve methods estimate cell reflectivities with the empirical expectation over the measurement distribution associated to the cell. Of course this yields poor estimates when the underlying distribution denoted as $P$ is far from the Gaussian distribution $Q$, as is often the case. Figure \ref{fig:histogramIntensity} illustrates this with aid of the Kullback-Leibler distance (KLD) denoted here as $ D_{KL}(P || Q)$. 

\subsection{Contribution and Outline}
\label{sec:contribution}

We propose to use an alternative grid representation of the ground which informs about reflectivity edges computed via fusion of individual reflectivity gradient grids from fixed laser-perspectives. We find that this representation achieves an invariance property to laser-perspective and vehicle motion which eliminates the requirement of applying additional post-factory laser reflectivity calibrations. The effectiveness of our formulation is supported through experimentation that demonstrates that: (a) our representation formulation is invariant to laser-perspective-motion, (b) this representation is effective for localizing in a prior-map using NMI registration of reflectivity edges at no additional computational burden.

The next section overviews state of the art laser reflectivity calibration methods used to generate reflectivity grids. In section \ref{Sec:ProposedApproach} we formulate our alternative reflectivity edges representation and establish our localization filtering framework. Section~\ref{Sec:Experiments} presents experimental results that validates localization with data collected by Ford's autonomous vehicles and Section~\ref{sec:conclusion} concludes our findings.

\section{Related Work}
\label{Sec:PriorArt}

Current applications involving usage of reflectivity based inference-grids use a prior after-factory reflectivity calibration stage to decouple measurements from the variations in reflectivity (e.g., those introduced by laser source, angle of incidence, range, etc). A good overview from the remote sensing community of these methods is available in~\cite{Kashani15}. Most of these approaches aim at estimating the response of each laser and its perspectives through an approximation curve fit between the measurement signatures of a reference target and its corresponding known reflectance. For example,~\cite{Ahokas.etal2006, Kaasalainen.etal2009, Pfeifer.etal2007, Pfeifer.etal2008} exploit standard calibration targets of known reflectance and uses least squares optimization to interpolate missing values and estimate the response of the laser scanner. Unfortunately, the options of target reflectances is very limited and measurements of these represent only a tiny fraction of the available data one may encounter in real world scenarios. Moreover, the cost of these targets is unfeasible for the general consumer. Follow up work in \cite{Khan16} explored a data driven approach to model laser intensities for surfaces with unknown reflectivity coefficient. This calibration process uses a standard white paper as a reference to estimate a relative measure of reflectivity. Although, this method is applicable even in the absence of standard materials with known reflectivity, it requires extrinsic parameter corrections by extracting the geometrical properties of the surfaces of different reflectivity.

Within the context of autonomous vehicles, \cite{Levinson10, Levinson.Thrun2010 } propose an automatic, target-less based calibration method that uses a probabilistic framework to learn the reflectivity response model of each laser from its measurements. This calibration is subsequently used for the purpose of localizing in a prior-map of the ground reflectivity.  The calibration method uses the expectation maximization iterative algorithm to generate the maximum likelihood model $p(y|x;b)$, that indicates the probability that for a given reflectivity $x$, laser $b$ observes reflectivity $y$. To establish correspondences, the algorithm iterates between improving both the likelihood of the surface reflectance $\Ybf$ and the laser responses. In practice, the whole calibration process is time consuming and computationally intensive since it requires the collection of map data and the reconstruction of a global reflectivity grid-map itself. Moreover, 
even application of state of the art calibration methods results in maps with low contrast, global smoothing, artifacts and noise. 
\section{Proposed Approach}
\label{Sec:ProposedApproach}


The application of our alternative representation framework is related to the online localization of an autonomous vehicle within a prior LIDAR map of the ground via registration corrections. Here, continuing the work of \cite{Castorena17} we propose edge alignments as our matching mechanism via corresponding isotropic gradients of the global and local grids. A similar matching criterion was envisioned in  \cite{Castorena.Kamilov.Boufounos2016} for the joint registration and LIDAR-camera fusion.

\subsection{Global prior-map}
\label{ssec:globalMap}

\begin{figure*}
	\centering
	\subfloat[Global Google map]{\includegraphics[trim=0cm 0cm 0cm 0cm, clip=true,width=1.0\linewidth, angle=0]{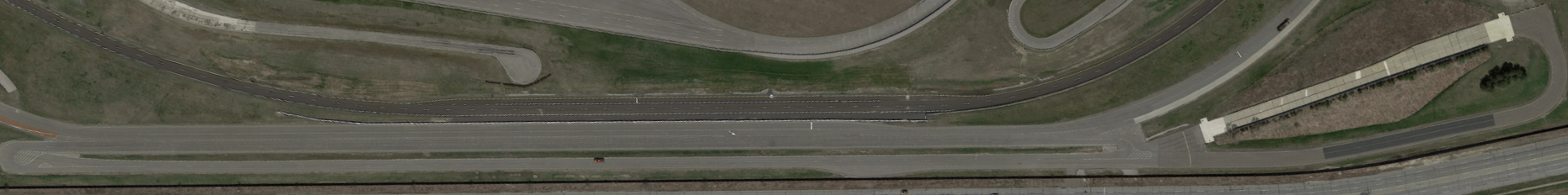}}
	
	\subfloat[Global LIDAR reflectivity map]{\includegraphics[trim=0cm 0cm 0cm 0cm, clip=true,width=1.0\linewidth, angle=0]{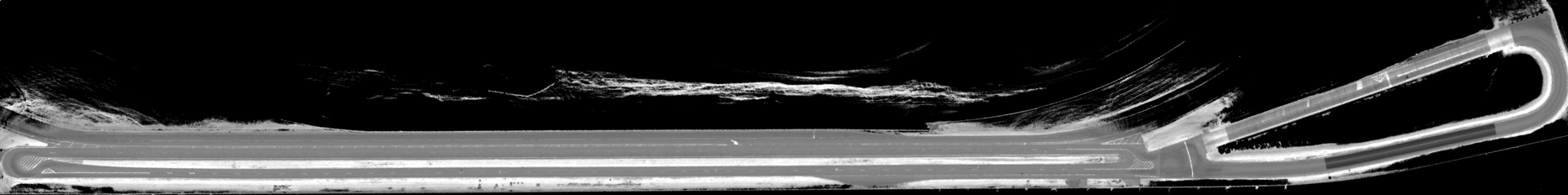}}
	
	\subfloat[Global LIDAR reflectivity edges map]{\includegraphics[trim=0cm 0cm 0cm 0cm, clip=true,width=1.0\linewidth, angle=0]{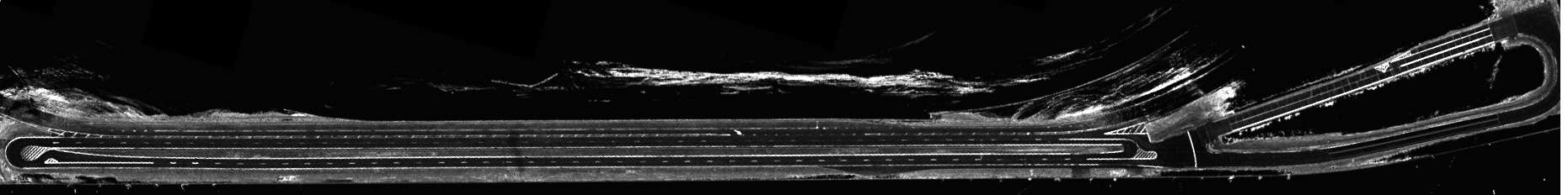}}	
	%
	\caption{Ortographic view of a global map of the ground obtained by surveying the region with our surveying vehicle.}
	\label{fig:globalMaps}
\end{figure*}

Estimation of the global prior-map modeling the environment uses the standard maximum a posteriori (MAP) estimate given the position measurements from GPS, odometry and 3-D LIDAR point-clouds of the entire trajectory \cite{Durrant-Whyte.Bailey2006, Bailey.Durrant-Whyte2006}. The MAP estimator is formulated as an equivalent least squares formulation with poses from a GPS with high uncertainty and contraints from both odometry and LIDAR scan matchings throughout the entire vehicle trajectory. Here, the LIDAR constraints are imposed via generalized iterative closest point (GICP) \cite{Segal.Haehnel.Thrun2009}. To minimize the least squares problem we use the implementation of incremental smoothing and mapping (iSAM) proposed in \cite{Kaess.Ranganathan.Dellaert2008} in an offline fashion. We also augment the GPS prior constraint with an artificial height prior (z = 0) to generate a planar graph. Once, we have the entire corrected pose trajectory we create a map by associating the 3D LIDAR points that fall within a ground height bound in the locality of the vehicle (i.e., in a local reference frame following the vehicle) with corresponding cells in a 2D grid. This grid is a discretization of the world along the x-y plane and each cell contains a reflectivity edge value characterizing the contained LIDAR reflectivity measurements in the cell and its neighborhood. A summary of the main steps involved in this process can be found in Algorithm \ref{poseMapAlgorithm}
\begin{algorithm}[t]
	\small 
	\caption{ {\small Pose + LIDAR to Ground Reflectivity-Edges Grid} } \label{poseMapAlgorithm}
	\begin{algorithmic}[1]
		\BState \textbf{Input:} Optimized pose $ G = \{ \Ybf_1, \Ybf_2,...,\Ybf_{\text{T}} \}$.
		\State \textbf{Set: }  Initialize 10 cm grid with cell values $[\d]_n \leftarrow \infty$
		\BState \textbf{for} $\Ybf_i$ in $G$ \textbf{do} 
		\State \quad // Extract ground point-cloud in local reference frame. 
		\State \quad // $\pmb{p}_j$ and $r_j$ are the metric point and its reflectivity, respectively. 
		\State \quad $\{ \{\pmb{p}_1,...,\pmb{p}_m\}, \{r_1,...,r_m\} \}$ = \text{ExtractGroundPoints( $\Ybf_i$)} 
		\State \quad \textbf{for} $j=0 \rightarrow m$ \textbf{do}		
		\State \qquad // Associate extracted ground points to 	
		grid (points in grid's
		\State \qquad // reference frame) and extract reflectivity edges.
		\State \qquad \text{Update:} $[\d]_n \leftarrow$  \text{ExtractReflectivityEdges( $\{\pmb{p}_j, r_j\} $ )}
		\State \quad \textbf{end for: }
		\State \textbf{end for: }
		\State \textbf{Output: } Grid map of ground reflectivity edges $ \d $
	\end{algorithmic}
\end{algorithm}
and the result of its application is shown through the example in Figure \ref{fig:globalMaps} including additional modalities.
We would like to emphasize that this 2D grid representation can not only represent 2D planar surfaces, rather it represents an ortographic view of a property of the world. For instance it has been used to represent a reflectivity of the ground as in \cite{Levinson.Thrun2007} or heights as in the work of \cite{Wolcott17}.

A global map is compactly denoted here by the reflectivity edges $ \dG \in \R^{N_x \times N_y} \cup \{ \infty\}^{N_x \times N_y} $ where $N_x$ and $N_y$ are the number of horizontal and vertical grid cells, respectively. The finite values in $\dG$ represent reflectivity edges while infinite values represent cells with insufficient information (e.g., neighboring reflectivities) to compute edges within the total of $N=N_xN_y$ cells. Note that in the remainder of this paper we often use $n\in\{1,\ldots,N\}$ to index the cells of $\d$ and other similarly sized matrices, essentially vectorizing them.

\subsection{Local-map}
\label{ssec:localMap}
\begin{figure}
	\centering
	\subfloat[Local LIDAR reflectivity map]{\includegraphics[width=0.3\linewidth]{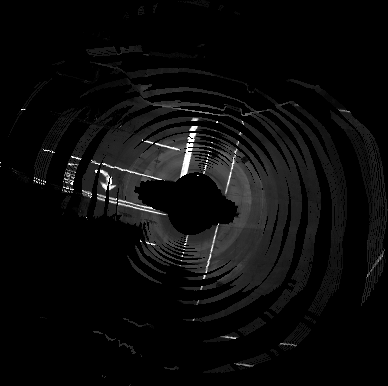}}
	\subfloat[Local LIDAR edges map]{\includegraphics[width=0.3\linewidth]{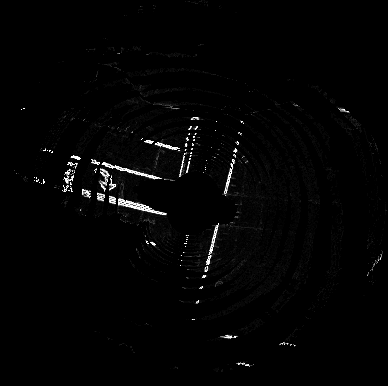}}
	\caption{Ortographic view of a local map of the ground of an area of size $40 \times 40$ m obtained from 8 full-LIDAR revolutions each at 10 Hz while undergoing vehicle motion.} 
	\label{fig:localmaps}
\end{figure}

The local map is also described with a 2D grid representing a discretization along the local x-y plane of the ground patch around the vehicle. The cell values within this local grid which we chose to be of size $40 \times 40$ m. (with a 10 cm cell resolution) represent the reflectivity edges computed and updated online from the accumulated LIDAR points which are motion compensated using inertial measurements and ground segmented also following the same process described in Section \ref{ssec:globalMap}. Overall this process is summarized in Algorithm \ref{poseMapAlgorithm}. The notation we use to describe the local map of reflectivity edges is $\dL \in \R^{M_x \times M_y} \cup \{ \infty\}^{M_x \times M_y}$ where finite and infinite values the same as in the global-map. An example that compares a local grid-map of calibrated reflectivities and of reflectivity edges is shown in Figure \ref{fig:localmaps}.
We would like to add here that there are sources that can make the number of cells with finite edge information in the grid to vary from scan to scan. These sources can include occlusions from the ego-vehicle and from other static and dynamic objects.

\subsection{Reflectivity Edges Inference}
\label{ssec:reflectivityEdges}

\begin{figure*}[htb]
	\centering
	\subfloat[Map-perspective of laser 1]{\includegraphics[width=0.25\linewidth]{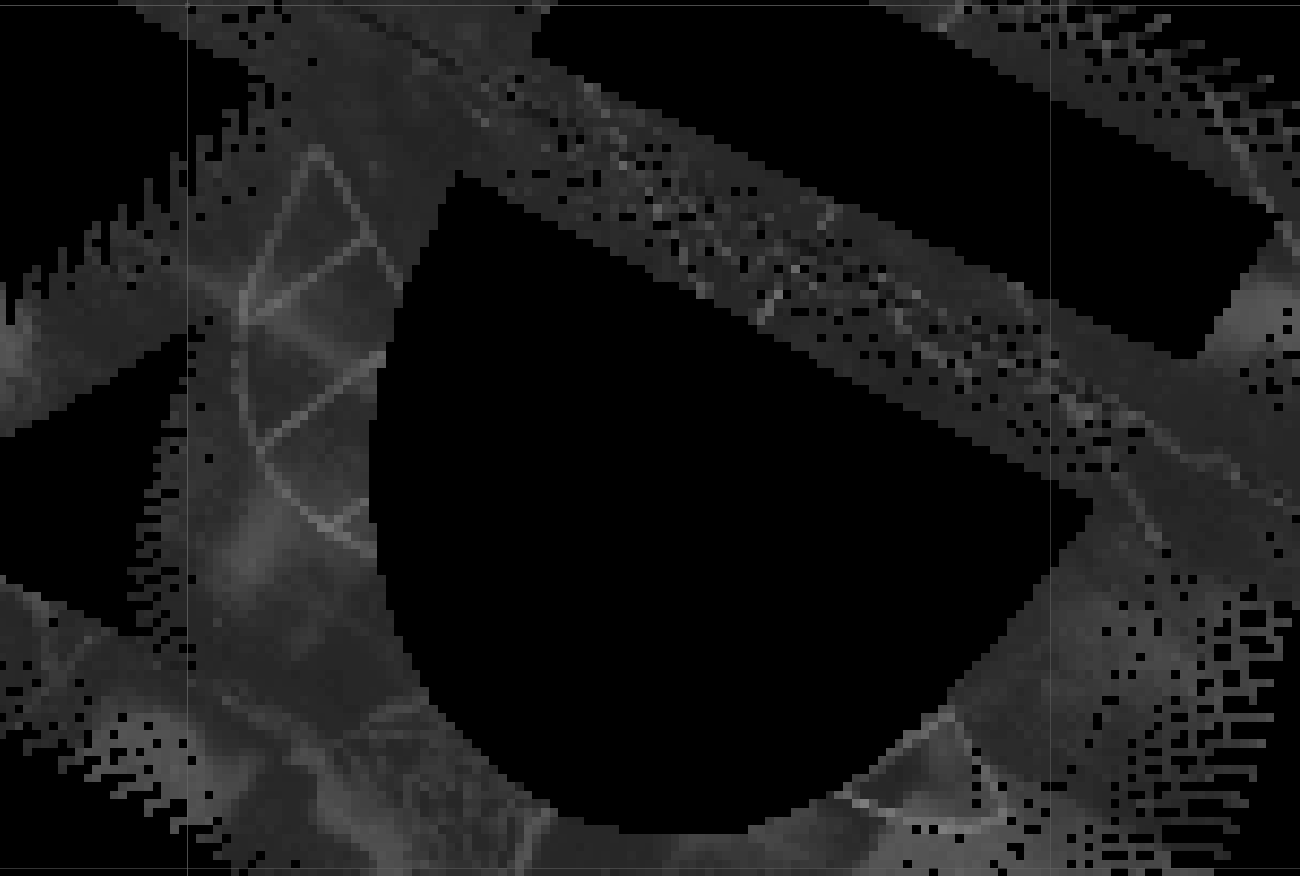}}
	\subfloat[Map-perspective of laser 16]{\includegraphics[width=0.25\linewidth]{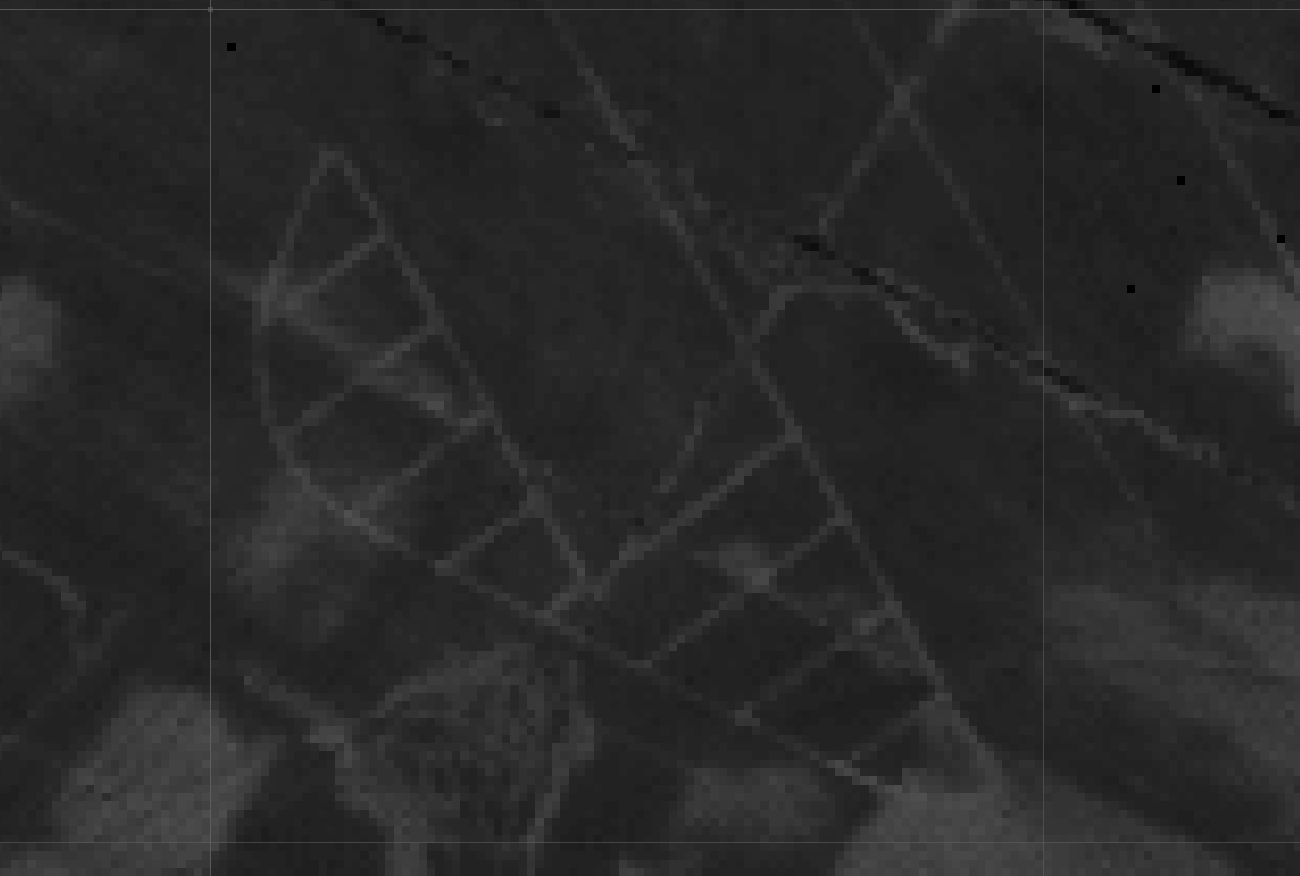}}
	\subfloat[Map-perspective of laser 19]{\includegraphics[width=0.25\linewidth]{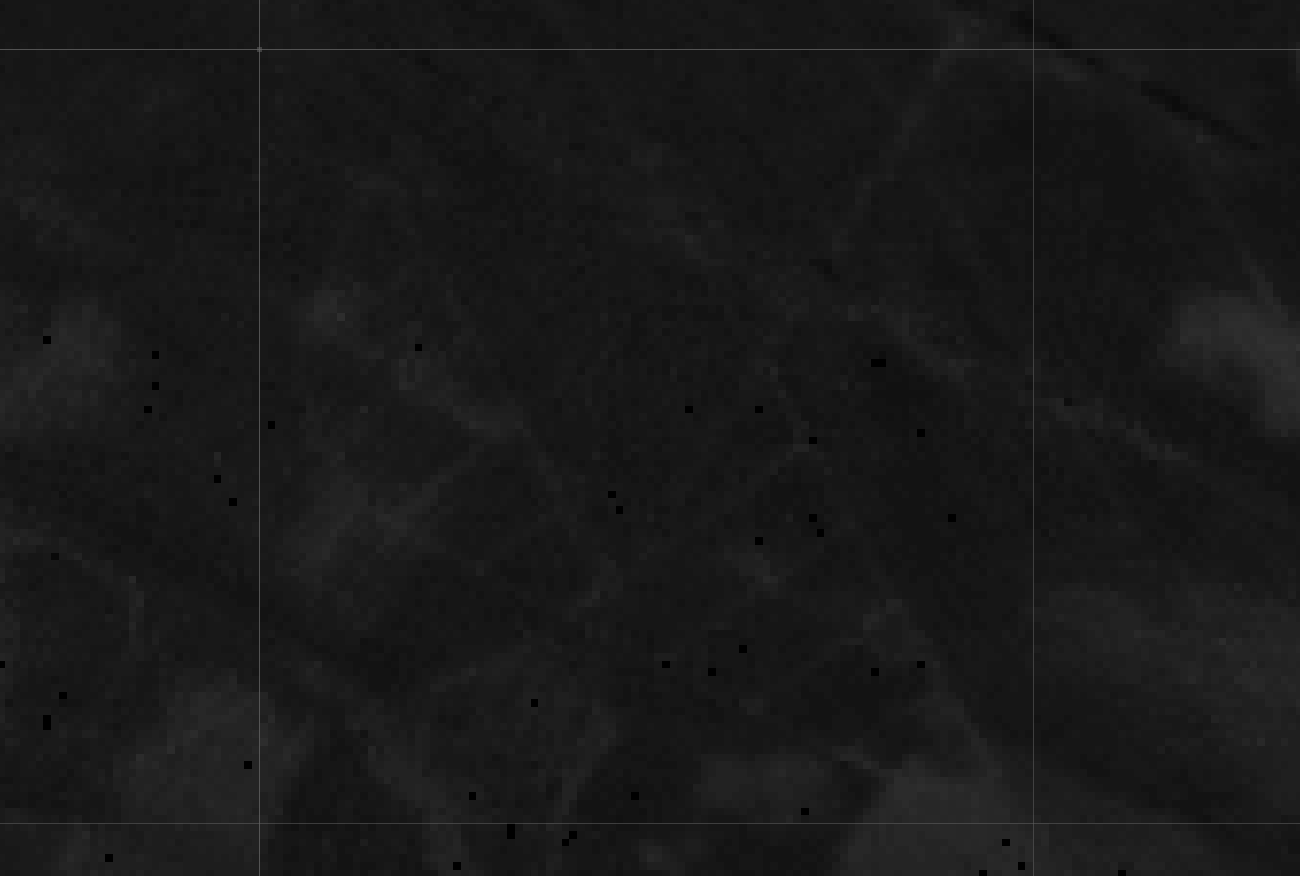}}
	\subfloat[Map-perspective of laser 21]{\includegraphics[width=0.25\linewidth]{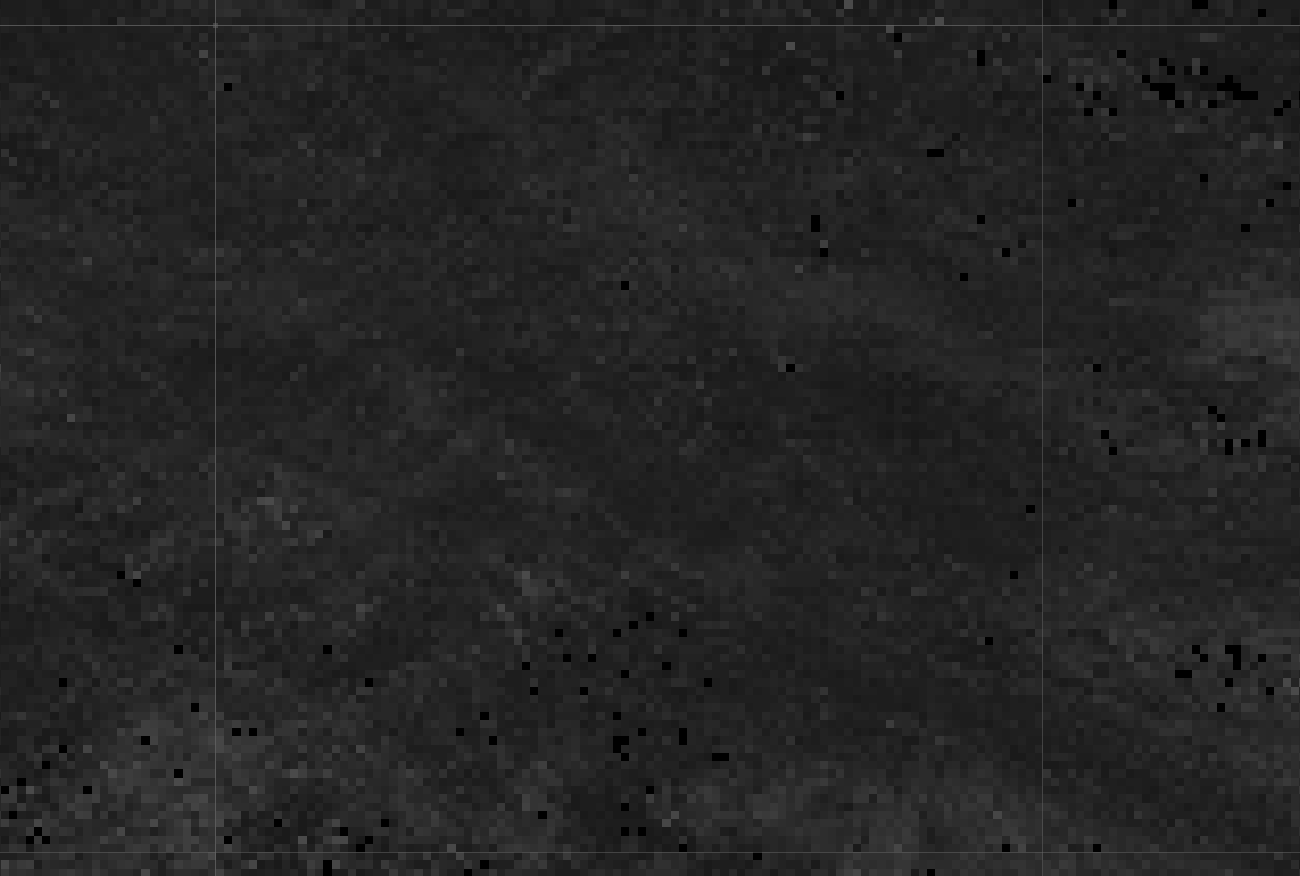}}
	\caption{Ortographic view of patches of the ground reflectivity from a fixed laser-perspective (indexed by order from bottom to top) in a Velodyne HDL-32E while surveying.  
		(a) View by laser 1 with restricted domain, (b) Full view by laser 16: reduced contrast, (c) View of laser 19: bad contrast from high range and angle of incidence, (d) View by laser 21: bad contrast.}
	\label{fig:reflectivityByBeam}
\end{figure*}

Reflectivity edges describe the grid-map regions with laser reflectivity discontinuities. These can be computed on the global or local grid representation using the reflectivity measurements associations to neighboring cells. Edges are computed here via the gradient field $ \g : \mathbb{R}^N \rightarrow \mathbb{R}^{N \times 2}$ defined as the first order finite forward difference along the horizontal  $ \g_x : \mathbb{R}^N \rightarrow \mathbb{R}^{N}$ and vertical $ \g_y : \mathbb{R}^N \rightarrow \mathbb{R}^{N}$ axes of the grid. These are defined point-wise as
\begin{equation} \label{Gradient}
[\gXbf ]_n = 
\begin{pmatrix} 
[\gxXbf]_n \\ 
[\gyXbf]_n 
\end{pmatrix} = 
\begin{pmatrix}
[\Xbf]_{n + N_y} - [\Xbf]_n \\
[\Xbf]_{n+1} - [\Xbf]_n
\end{pmatrix}.	
\end{equation} 
Such defintion is popular in total variation (TV) image denoising \cite{Rudin92, Chambolle04} because of its stability and isotropy. Other definitions can be used in place as long as the support of the filter remains small and the gradient is computed within a close neighborhood to reduce the likelihood of cells with insufficient information to compute edges.

\begin{figure}[htb]
	\centering
	{\includegraphics[width = 9 cm]{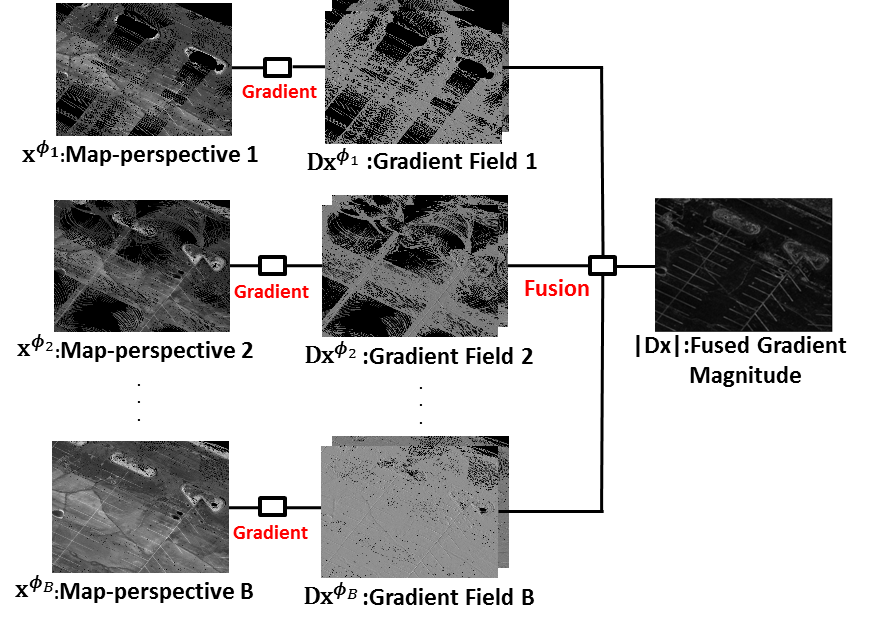}}
	\caption{Schematic of our inference-grid representation formulation. The input images in the left column represent individual laser grids obtained from \eqref{perspectives}. The gradient field for each of these is then computed in the second column and fused.}
	\label{fig:schematic}
\end{figure}

Since there are dependencies of the reflectivity measurements on the laser source, angle of incidence and range, careful consideration on how to compute the gradient should be accounted to avoid capturing these variations. An illustration of these dependencies can be observed in Figure \ref{fig:reflectivityByBeam}. 
Here, we propose to address this by computing edges as a fusion of individual laser-perspective gradient grids as illustrated in Figure \ref{fig:schematic}. 
%
Individual gradient grids $ \Dbf \Xphibf $ represent the edge contribution of all the reflectivity measurements from a fixed set of laser-perspective parameters $\phibf$ where  for example, $\phibf = (b, \theta, r)$ indicates the laser with global index $b$ at an angle of incidence $\theta $ and range $r$ and where we assume $ B = |\Phibf| < \infty$ is finite with $\phibf \in \Phibf$ and $\Phibf = \{ \phibf_1, \phibf_2,...\phibf_B \}$. The individual laser-perspective $ \Xphibf \in \R^{N_x \times N_y} \cup \{ \infty\}^{N_x \times N_y} $ representing the ground reflectivity from the view of $\phibf$ is cell-wise estimated via the empirical expectation
\begin{equation} \label{perspectives}
	[\Xphibf]_n = 
	\begin{cases} 
		\frac{1}{\Mphin} \sum\limits_{m = 1}^{\Mphin} [\yphin]_m, & \mbox{for} \quad n \in \Omega^{\phibf} \\
		\quad\quad\quad \infty & \mbox{for } \quad n \notin \Omega^{\phibf}
	\end{cases}
\end{equation}
where $\yphin \in \R^{M_n^{\phibf}}$ are the $ {M_n^{\phibf}} $ reflectivity measurements associated with cell $n$.

The fusion stage of individual gradient grids consist in the average overall available $\Xphibf$ (with $\infty$ values ignored as in \eqref{equivalence}). In other words, fusion is carried via 
\begin{equation} \label{fusion}
	\widehat{\g \Xbf} = \frac{1}{|\Phibf|} \sum_{\phibf \in \Phibf} \g \Xphibf.
\end{equation}
where $\widehat{\g \Xbf} \in \mathbb{R}^{N \times 2}$ denotes the fusion estimate and the fused grid domain $\O' \subset \R^2$ is given by the union $	\O' = \bigcup_{\phibf \in \Phibf} \Ophi  $ of the individual laser-perspective domains $\Omega^{\phibf}$.
Note that by generating the individual laser-perspective grids a normalization that unbiases the non-uniform number of measurements from a laser and its perspectives in cells is carried out. Under this representation, the gradients of reflectivity inference grid components $[\g \Xphibf]_n$ become statistically stationary which allows one to operate on each invariantly. We would like to add here that for ease of notation, every element-wise operation that involves an $\infty$ value will be determined by ignoring it from its computation. For example,  
\begin{equation} \label{equivalence}
\sum\limits_{\phibf \in \Phibf} [\Xphibf]_n = \sum\limits_{\phibf \in \Phibf} [\Xphibf]_n \cdot \textbf{1} \{ [\Xphibf]_n < \infty \}.
\end{equation}

\subsection{Normalized Mutual Information for Registration}
\label{ssec:registration}

The planar GPS constraint of zero height in the optimization of the global prior-map described in section \ref{ssec:globalMap} simplifies the registration problem to a 3-DOF search over the longitudinal (x), lateral (y) and head rotation (h) vehicle pose. Here, we propose to maximize the normalized mutual information (NMI) \cite{Studholme1999} between reflectivity edges of candidate patches $\d_{\text{G}|\Omega_{\tbf}} = | \g\X_{\text{G}}|_{|\Omega_{\tbf}} $ from the prior map and the local grid-map $\dL = | \g \Xbf_{\text{L}} |$. In other words, 
\begin{equation} \label{NMI}
	\widehat{\tbf} = \arg \max\limits_{ \tbf \in \Tcal } 
					\left \{  
						\text{NMI}(\dL, \d_{\text{G}|\Omega_{\tbf}} )
					\right \},
\end{equation}
where $\Tcal$ is a discrete neighborhood set centered at the (x,y,h) position and orientation and
\begin{equation}
	\text{NMI} (A, B ) = \frac{ H(A) + H(B) }{H(A,B)}.
\end{equation}
 $H(A)$ and $H(B)$ is the entropy of random variable $A$ and $B$, respectively and $H(A,B)$ is the joint entropy. Using NMI over standard mutual information (MI) has the advantages of being robust under occlusions \cite{Dame11}, less sensitive to the amount of overlap \cite{Studholme1999} while also maintaining the desirable properties of MI registration including outlier robustness \cite{Viola95}.

\subsection{Online Localization Filter}
\label{ssec:localization}

The localization framework consists of two main steps: a NMI registration search based on \eqref{NMI} and a fusion with inertial measurements from the IMU. The procedure used here follows that described in \cite{Wolcott14} fusing via an extended Kalman filter (EKF) the result of the registration with the inertial measurements. Mathematically, this fusion is performed via the iterative localization updates
 \begin{alignat}{2} \label{EKF}
	\text{Predict:} & \quad \bar{\tbf}_k = F_{k-1} \tbf_{k-1} \\
	& \quad \bar{\Sigma}_k = F_{k-1} \Sigma_{k-1} F_{k-1}^T + Q_{k-1} \nonumber \\
	\nonumber \text{Update:} & \quad K_k = \bar{\Sigma}_k H^T_k ( H_k \bar{\Sigma}_k H^T_k + R_k)^{-1} \\
	\nonumber & \quad \tbf_k =  \bar{\tbf}_k + K_k(z_k - h_k( \bar{\tbf}_k )) \\
	\nonumber & \Sigma_k = (I - K_k H_k) \bar{\Sigma}_k (I - K_k H_k)^T + K_k R_k K_k^T
 \end{alignat}
where $F_k$ is the plant model integrating measurements from the IMU and $Q_k$ is the corresponding uncertainty, $z_k$ is the output of the NMI registration in \eqref{NMI} and $R_k$ is the corresponding uncertainty estimated as a fit to the covariance of the NMI cost as was done in \cite{Olson09}. The filter in \eqref{EKF} is initialized in a global reference frame with a GPS measurement providing an initial pose and orientation guess with high uncertainty. The NMI registration stage is performed with a dynamic bound exhaustive search adaptively updated to a 3$\sigma$ window around the posterior distribution of the EKF. This exhaustive search prevents our registration to fall within the local minima that might be present in the NMI cost function.
We would like to mention here that the finest search is constrained to be of resolution of 20 cm over a \textpm1 m window and that our NMI of reflectivity edges is agnostic to the filtering approach and thus other methods could also be applied instead.

\section{Experiments}
\label{Sec:Experiments}

\begin{figure*}[!t]
	\centering
	\subfloat[
	$D_{KL}(P || Q) = 0.33 $ bits]{ {\includegraphics[width = 5 cm]{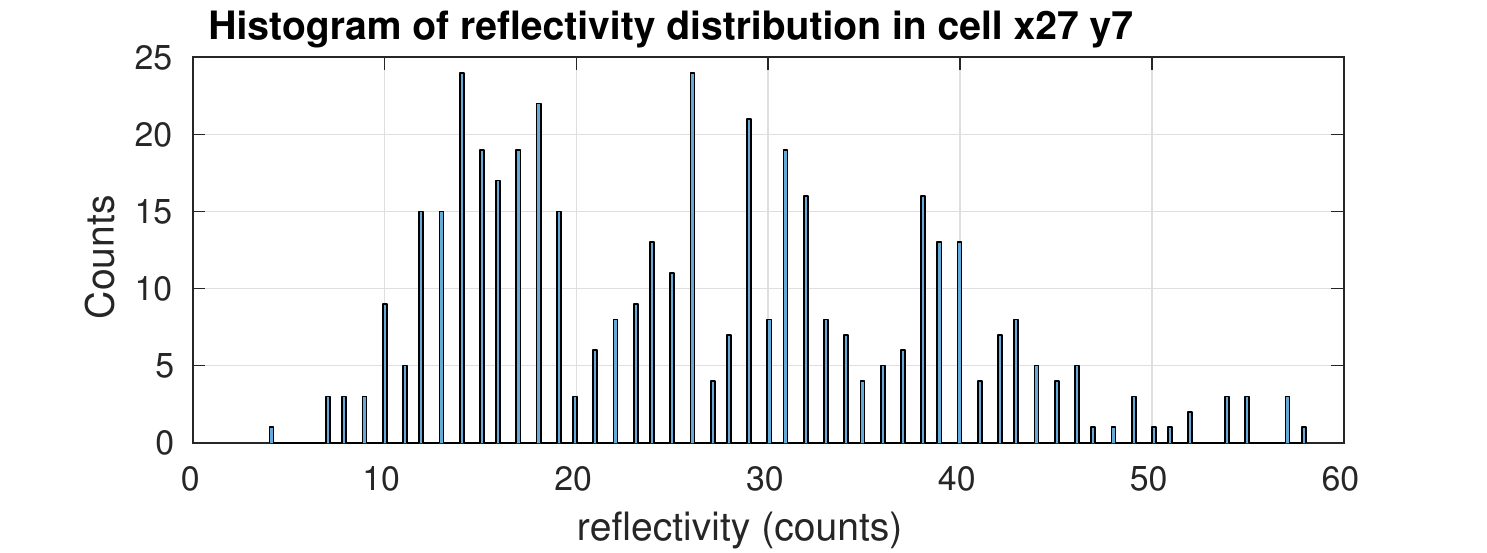} } }
	\subfloat[ 
	$D_{KL}(P || Q) = 0.82 $ bits]{ {\includegraphics[width = 5 cm]{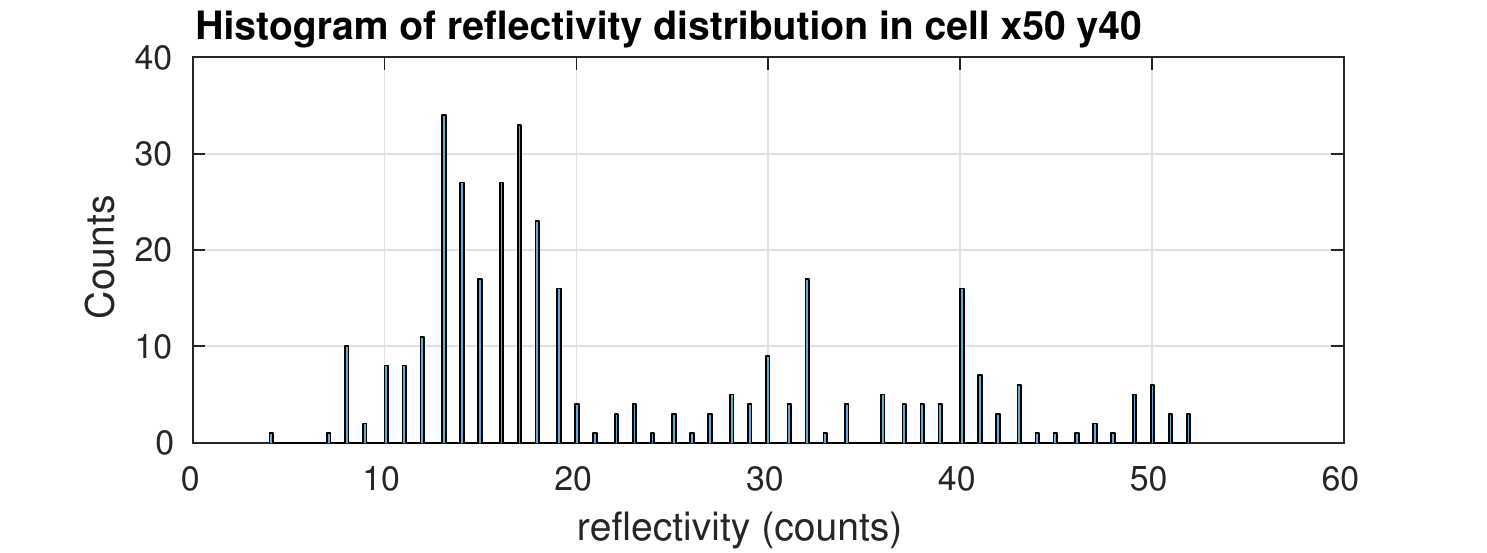} } }
	\subfloat[ 
	$D_{KL}(P || Q) = 0.22 $ bits]{ {\includegraphics[width = 5 cm]{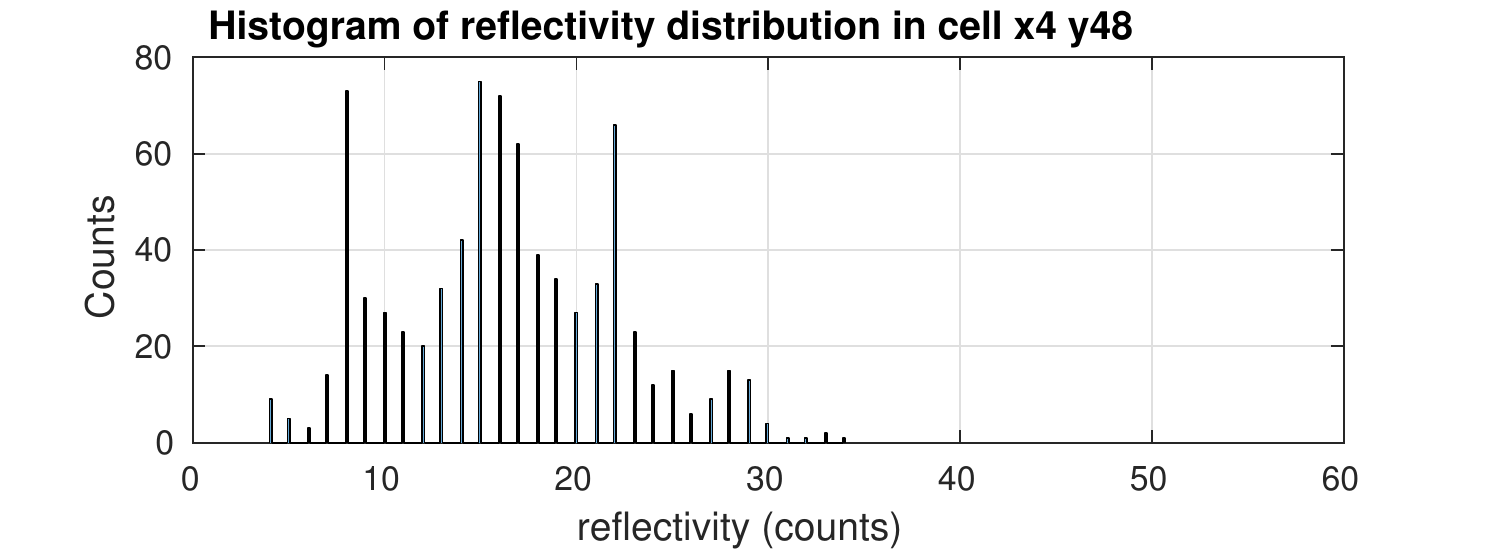} } }

	\subfloat[ 
	$D_{KL}(P || Q) = 0.09 $ bits]{ {\includegraphics[width = 5 cm]{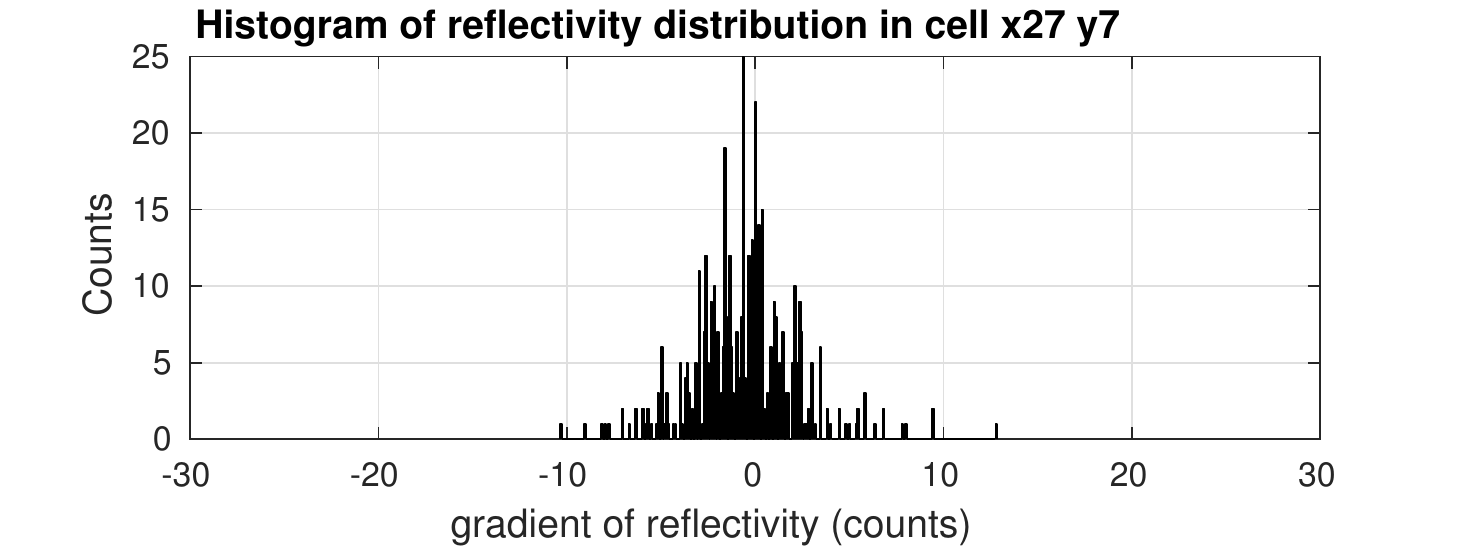}} }
	\subfloat[ 
	$D_{KL}(P || Q) = 0.15 $ bits]{ {\includegraphics[width = 5 cm]{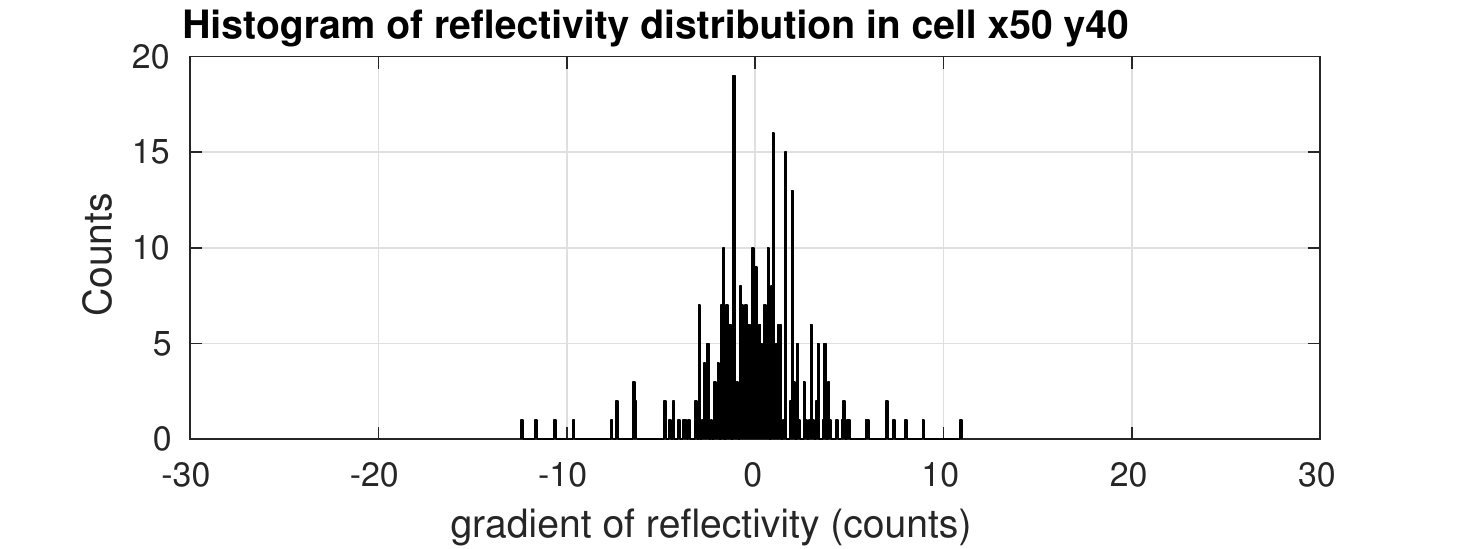} } }	
	\subfloat[ 
	$D_{KL}(P || Q) = 0.02 $ bits]{ {\includegraphics[width = 5 cm]{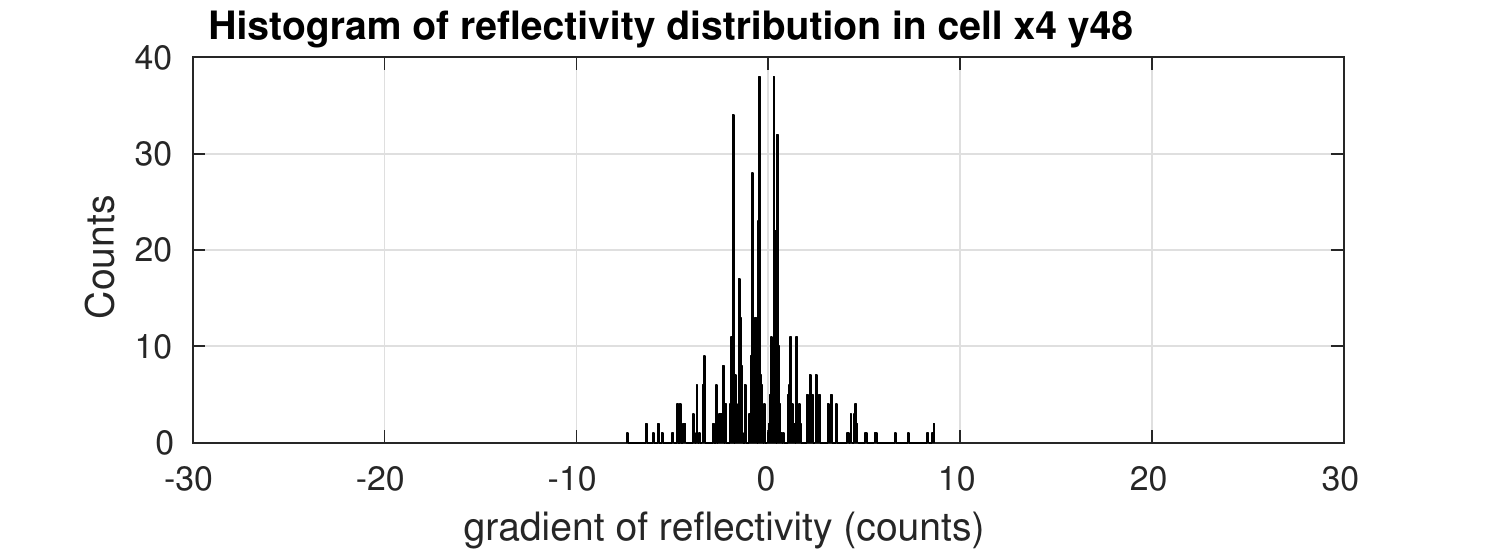}} }
	\caption{Histogram distributions of measurements from multiple HDL-32E's in $10 \times 10$ cm. cells. (a-c) Raw reflectivities. (d-f) Gradient of reflectivities.}
	\label{fig:histogramReflectivityComparison}
\end{figure*}
%

To validate our approach, we use log datasets 
collected with Ford’s Fusion testing fleet of autonomous vehicles at the Ford campus and downtown area in Dearborn, Michigan. These vehicles are outfitted with four Velodyne HDL-32E 3D-LIDAR scanners and an Applanix POS-LV 420 (IMU). All four HDL-32E’s were mounted on the roof above the B pillars of the vehicle; two which rotate in an axis perpendicular to the ground (i.e., non-canted) and two that are canted. Extrinsic calibration of the LIDAR’s is performed using GICP \cite{Segal.Haehnel.Thrun2009}. 
%
To test our approach, we use data only from the two non-canted LIDARs. Under this configuration, the angle of incidence and range per laser recorded from the ground remains approximately constant over the course of the 360\textdegree scan trip. Thus, each laser index implicitly accounts for these two which then results in a total of $B = 64$ laser-perspective parameters.
%
To compare against state of the art on ground laser reflectivity based localization, we test against the method of \cite{Levinson10,Levinson.Thrun2010} involving an implementation of a deterministic reflectivity calibration process. The main difference in our implementation is that we make an approximation to generate the look-up table in one data pass. We would like to mention that the calibration and map generation dataset is the same thus ensuring the best possible map results with this method. To evaluate localization performance, we perform the NMI registration based filtering localization approach described in Section \ref{ssec:localization} and compare it against (full) bundle adjustment pose. In the absence of known ground-truth, this method has been used in the literature as an alternative to obtain a quantitative measure of error \cite{Levinson10, Wolcott14, Wolcott17}. In essence, this error compares the performance of the online localization approach to the offline (full) bundle adjustment trajectory correction one would obtain with GPS/IMU and 3-D LIDAR scan-to-scan matching constraints. Our implementations were all done in C on a Dell Precision 7710 and tested in our autonomous testing fleet. 

In the first part of this experimentation we present illustrations of cell reflectivity distributions and metric results of the Kullback-Leibler distance (KLD) along cells in a grid. The idea here is that the KLD will measure how much the distribution of reflectivities deviates from a corresponding Gaussian model. Since inference grid based methods estimate a reflectivity value from its distribution, we use the KLD to asses the quality of a calibration or a representation transformation to achieve the whitening of its data. Figure \ref{fig:histogramReflectivityComparison} includes in (a)-(c) the distribution of raw reflectivities in individual cells whereas (d)-(f) includes the corresponding distributions of the gradient representation. In addition, we also include Table \ref{tab:table1} 
\begin{table}
	\caption{\label{tab:table1} Kullback-Leibler distance}
	\centering
	\begin{tabular}{lccc}
		\hline
		\\
		Road scene category & Raw & Levinson and Thrun \cite{Levinson.Thrun2010} & Ours \\
		\hline
		\\
		Road Markings & 0.2383 & 0.2243 & \textbf{0.1481} \\
		Asphalt & 0.2714 & 0.2724 & \textbf{0.1438} \\
		Ice patches & 0.3647 & 0.2334 & \textbf{0.1097}  \\
		Overall & 0.2860 & 0.2644 & \textbf{0.1387}  \\
		\hline
	\end{tabular}
\end{table}
which includes a comparison of averaged KLD's over occupied cells in road regions of multiple categories for the three methods: raw reflectivity, Levinson \& Thrun's calibration \cite{Levinson10, Levinson.Thrun2010} and ours. These regions were predefined by segmenting the road into the different reflectivity class materials. Results show that in every case the proposed edge representation substantially whitens or decouples reflectivity measurements more than state of the art calibration based methods thus being more effective in achieving invariance to laser source, perspective and vehicle motion. 

The rest of this manuscript is dedicated to demonstrate the localization capabilities of our proposed approach. The environment on which localization performance is assessed includes a variety of interesting scenarios. These scenarios involve the presence of non-uniform ground surfaces, less structured roadside surfaces and occlusions. To detail the behavior of our framework under these circumstances we include localization performances in small log snippets that include for the most part the specific cases of the aforementioned examples. Table \ref{tab:table2} summarizes the localization performance comparison reporting the longitudinal, lateral and head angle RMSE obtained when driving over the corresponding specific cases while a more detailed discussion is included thereafter.
\begin{table*}
	\caption{\label{tab:table2} RMSE Localization performance comparison}
	\resizebox{\columnwidth}{!}{%
	\centering
	\begin{tabular}{lcccccc}
		\hline
		\\
		\multirow{2}{*}{Road Scene Category} &\multicolumn{3}{c}{Levinson and Thrun \cite{Levinson.Thrun2010}} & \multicolumn{3}{c}{Ours} \\
		& Longitudinal & Lateral & Head & Longitudinal & Lateral & Head \\
		& (cm) & (cm) & (rads)& (cm) & (cm) & (rads)\\
		\hline
		\\
		a) Urban marked (UM) with dynamic occlusions & 8.3 & 2.7 & 2e-3 & 7.8 & 2.2 & 2.3e-3\\
		b) Bridge ramp, UM & 6.9 & 1.9 & 3e-3 & 4.7 & 1.55 & 2.5e-3\\
		c) Roadside Grass, UM  & 8.9 & 5.0 & 6e-3 & 2.8 & 2.5 & 7e-3\\
		d) Straight road, UM & 7.7 & 1.4 & 1.7e-3 & 4.5 & 1.4 & 1.4e-3 \\
		e) Curvy road, urban multiple marked lanes (UMM) & 6.4 & 3.2 & 2.6e-3 & 3.3 & 1.7 & 2.4e-3 \\
		\hline
	\end{tabular}
}
\end{table*}
 Note that in both methods the longitudinal error is higher than the lateral due to the presence of fewer road features to constraint in the longitudinal direction as has been also corroborated in \cite{Levinson10, Wolcott14}. Also note that in general, our method presents slightly better localization performance in comparison to Levinson's. This is mainly due to the improved preservation of edges as opposed to \cite{Levinson10} whose method can be interpreted as a global smoothing of the map losing edge sharpness and thus registration accuracy.

\begin{figure}[htb]
	\centering
	\subfloat[Occlusion in point-cloud]{\includegraphics[width=0.3\linewidth]{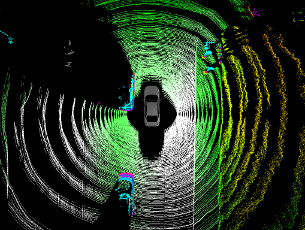}}
	\subfloat[Occlusion in reflectivity edges map]{\includegraphics[trim=0cm 2.345cm 0cm 2.77cm, clip=true, width=0.3\linewidth]{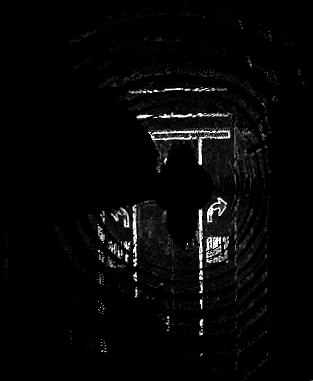}}
	\caption{Ortographic view of occlusion in local LIDAR perception. a) LIDAR point cloud accumulation (color coded according to height), b) Reflectivity edges local map.}
	\label{fig:occlusion}
\end{figure}

The first case we present relates to the effect of occlusions mainly caused in our datasets by the presence of static objects (e.g., trees, traffic signs, buildings, walls, barriers), dynamic objects (e.g, bicycles, vehicles) and the ego-vehicle occluding the FOV. Our observations confirmed that the presence of occlusions whose corresponding points were mostly removed in the segmentation of points that belong to the ground caused a reduction of edge information in the grid-maps. In the construction of the global map this effect was reduced by the multiple views induced by motion and by multiple passes through the same regions. The local grid of reflectivity edges however, presented a reduction of reflectivity edge information in the presence of occlusions as illustrated in Figure \ref{fig:occlusion} which shows two vehicles, traffic signs and the sidewalks. 

The results of localization on the snippet log around Downtown Dearborn presenting these occlusions is summarized in Table \ref{tab:table2} category "a" which also makes a comparison against \cite{Levinson10}. Note that in this case the longitudinal errors are the second highest in comparison to other road scene categories mainly due to a loss of the field of view (FOV) in the presence of occlusion. However, these are still within $\pm$10 cm which supports previous findings that NMI is less sensitive to the amount of overlap and robust against occlusions \cite{Studholme1999, Dame11}. When most of the ground is occluded the localization uncertainty is limited to that of inertial measurements due to the absence of reflectivity edge information for registration. 

\begin{figure*}[htb]
	\centering
	\subfloat[Global map section from Google maps]{\includegraphics[width=0.5\linewidth]{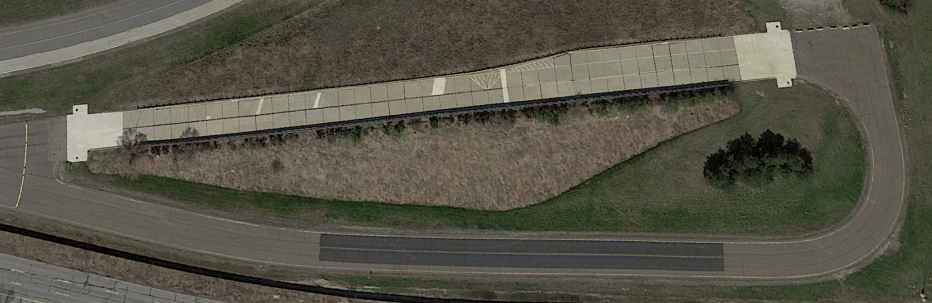}}
	\subfloat[Global map section showing 3D ground points of bridge]{\includegraphics[trim=0cm 4.8cm 0cm 4cm, clip=true, width =0.5\linewidth]{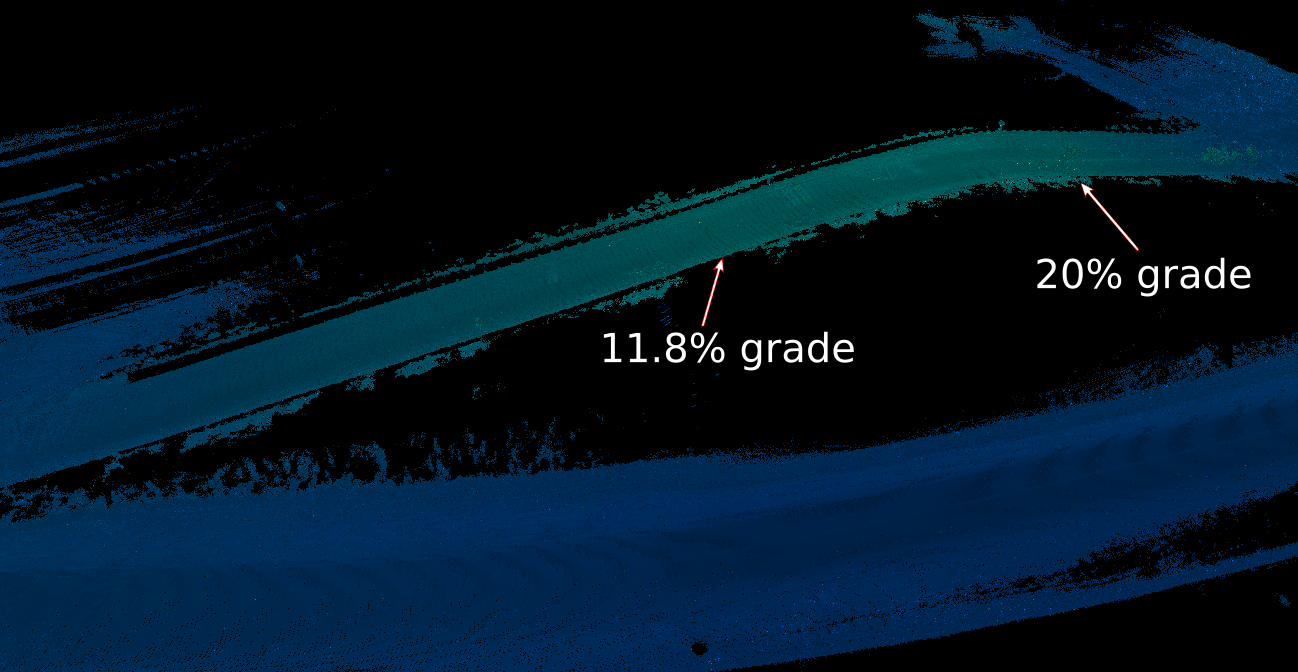}}
	
	\subfloat[Global map section from LIDAR reflectivity]{\includegraphics[trim=0cm 0cm 0cm 0.29cm, clip=true, width =0.5\linewidth]{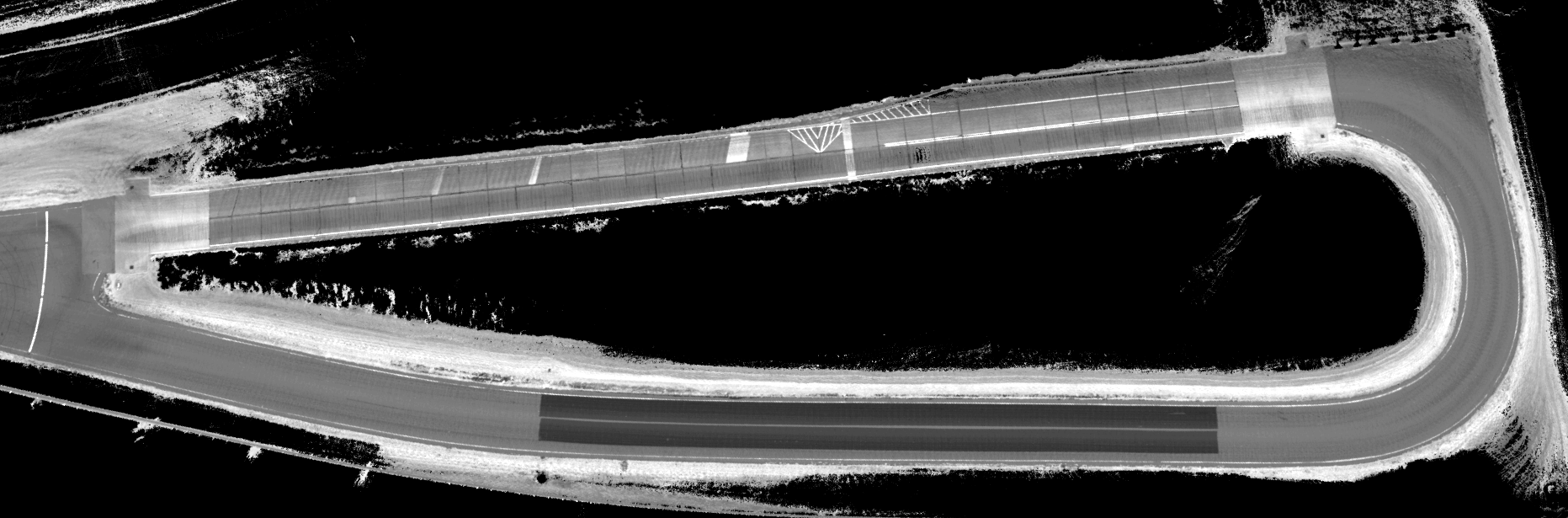}}
	\subfloat[Global map section from LIDAR reflectivity edges]{\includegraphics[width =0.5\linewidth]{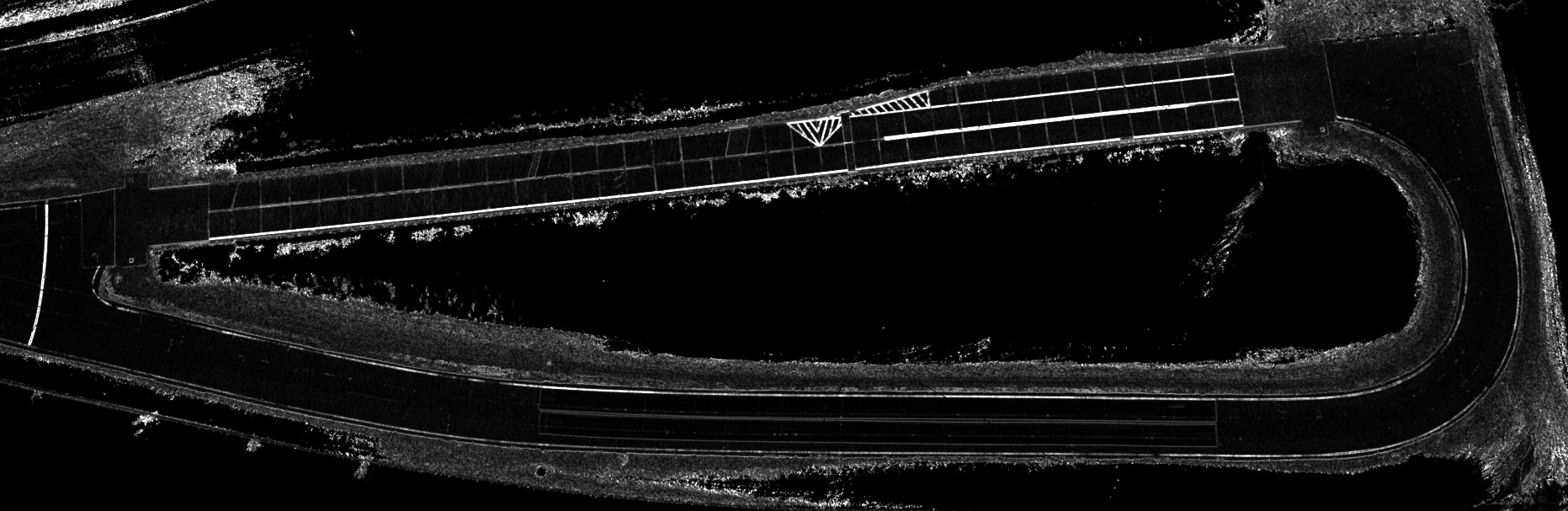}}
	
	\caption{Ortographic view of a ramped bridge section of a global map from different modalities.}
	\label{fig:rampMap}
\end{figure*}

In the second case we present the effect of non-uniform surfaces on localization performance. We tested this in the bridge shown in Figure \ref{fig:rampMap} including two slopes of $11.8\%$ and $20\%$ grade by  driving in both directions. 
Note that our approach well captures scene reflectivity edges even in the presence of non-uniform road surfaces. In the case of generating the local grid map we found that sufficient edge information results as long as the road is smooth (in the local region around the vehicle) and regardless of its height. The results of localization shown in Table \ref{tab:table2} category ``b" summarize and indicate slightly better performance than \cite{Levinson10}. This shows that although there are variations in the amount of overlap in sequential updates of the local map, our NMI based approach is robust against these variations. The low lateral error for this category is due to the higher constraints imposed by the presence of straight road lane markings in this direction. This is also in agreement with the lateral errors found in the straight road category d). The low longitudinal error in b) is due to the presence of road features in this direction as illustrated in the section of Figure \ref{fig:rampMap} corresponding to the bridge. As a note we would like to add that we were not able to find a drivable scenario in Dearborn and in Ford's campus in which the non-uniformity of the road was rough enough to significantly reduce the reflectivity edge information to a level where registration would not reduce the uncertainty below that of inertial measurements.

In the third case, we evaluated the localization performance in regions with less structured surfaces in the roadside (e.g., grass). We found that the segmentation process that extracts ground points for both the global and local grid map generation automatically removed most of the roadside regions since these were above the sidewalk outside the ground height bound. One thing worth mentioning is that our bounded range restriction (e.g., ~20 meters) for LIDAR points further prevents these regions to be within the FOV. There were two regions of straight and curvy road which we isolated into a log snippet where a significant portion of roadside grass remained even after the segmentation step. These regions yielded reflectivity measurements with high uncertanties due to several reasons. For instance, a Lambertian surface model can no longer be assumed and the statistics of these regions can significantly change over time due to for example grass growth or change of season. We have included in Table \ref{tab:table2} category ``c" the RMSE localization performance results comparing with those from Levinson \cite{Levinson10}. Note that the localization performance in these regions is slightly better than those produced by Levinson's method utilizing absolute calibrated reflectivity values. The reason for the slight increase on the (h) error in both methods compared to other road categories is that we drove in circles and thus the head angle was less constrained than when driving straight. We would like to add to this test case that most of the data logs on which we tested performance contained roadside grass, rocks, bushes and other non-structured surfaces thus showing robustness to these scenarios. 

		
		
		

\begin{figure}
	\centering
	\subfloat[Longitudinal error]{\includegraphics[width=0.5\linewidth]{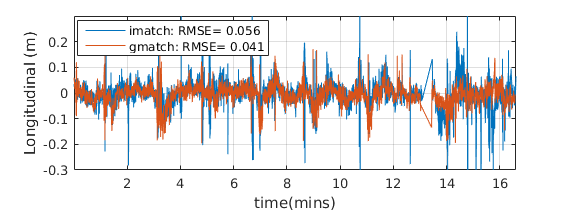}}
	%
	\subfloat[Lateral error]{\includegraphics[width=0.5\linewidth]{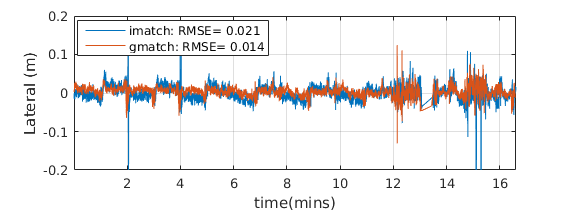}}
	
	\subfloat[Head angle error]{\includegraphics[width=0.5\linewidth]{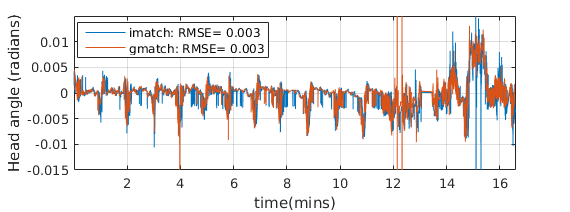}}
	
	\caption{Localization residual error comparisons.}
	\label{fig:performance}
\end{figure}

Finally, we present the localization performance results in a full route which includes crossing multiple times through the bridge in Figure \ref{fig:rampMap}, grass on the roadside and occlusions from trees, traffic signs, cones placed in the road, barriers, walls, the bridge itself and some vehicles. We drove along this route for a period of approximately $17$ mins and performed over 8500 NMI registrations.  The localization results throughout the entire trajectory are summarized in Figure \ref{fig:performance} and compared with the method of \cite{Levinson10} for the longitudinal (x), lateral (y) and head angle (h). In this Figure, the plot with legend ``gmatch" represents the performance of our reflectivity edge based matching while ``imatch" represents that of our implementation of \cite{Levinson10}. Here, we found that an RMSE of (x) 4.1 cm, (y) 1.4 cm and (h) 2.5$e-3$ rads were achieved with our proposed method versus the corresponding 5.6 cm, 2.1 cm and 2.5$e-3$ that resulted from \cite{Levinson10}. In addition to the general higher longitudinal error, Figure \ref{fig:rampMap}.a presents a larger number of spikes than \ref{fig:rampMap}.b. The reason for sudden jumps in error are the miss-registrations from poor features to constraint the longitudinal direction which is common in roads with straight lane markings. Fortunately, even these spikes are bounded within  $\pm$20 cm with our proposed method. The higher frequency of spikes throughout all Figures \ref{fig:rampMap}.a-c are caused from the ego-vehicle coming to full stop for a few seconds on multiple times which sparsifies the local-map. One additional advantage of our procedure is that a post-factory reflectivity calibration process to reduce the variations in response across the multiple lasers observing the ground is not required to compute any of the global or local grid maps. This process can currently take up to 4 hrs to a skilled artisan per vehicle which is prohibitive when production of these scales up to the masses.
\section{Conclusion}
\label{sec:conclusion}

In this investigation we demonstrate a representation of LIDAR reflectivity grid maps that is invariant to laser, perspective and vehicle motion. In particular, we validate the effectiveness of this representation in robotic applications involving localization within a prior grid-map of the ground. 
Our new approach is computationally tractable in real time and removes the requirement of any post-factory reflectivity calibration while achieving slightly better performance than state of the art on LIDAR reflectivity ground based grid localization. This represents a significant advantage if one considers the cost implied in calibration against standard reference targets and/or the time requirements of these processes, which are currently unfeasible for robots/vehicles produced in mass.




\bibliographystyle{files/IEEEbib}
\bibliography{files/refs}

\end{document}